\documentclass{article}
\usepackage{times, wrapfig, epsfig, a4, theorem, amssymb, amsmath, setspace}
\usepackage{graphicx}
\usepackage{url}
\usepackage{algorithm, algorithmic}
\usepackage{array}

\usepackage{natbib}

\usepackage{graphicx} 

\usepackage{natbib}

\usepackage{algorithm}
\usepackage{algorithmic}

\begin{document}
\title{Building Diversified Multiple Trees for Classification in High Dimensional Noisy Biomedical Data}
\author{Jiuyong Li, Lin Liu, Jixue Liu and Ryan Green \\ School of Information Technology and Mathematical Sciences \\ University of South Australia, Australia}
\date{}
\maketitle

\begin{abstract}
It is common that a trained classification model is applied to the operating data that is deviated from the training data because of noise. This paper demonstrates that an ensemble classifier, Diversified Multiple Tree (DMT), is more robust in classifying noisy data than other widely used ensemble methods. DMT is tested on three real world biomedical data sets from different laboratories in comparison with four benchmark ensemble classifiers. Experimental results show that DMT is significantly more accurate than other benchmark ensemble classifiers on noisy test data. We also discuss a limitation of DMT and its possible variations.
\end{abstract}


\section{Introduction}
Classification aims at building models with labeled data, and
applying the models to unlabelled data to predict their labels. For example, a fraud detection system is built on known
fraudulent cases and normal cases, and is used to predict coming fraudulent cases. A
fundamental assumption is that the system operating environment is
very similar to the system building environment. Unfortunately, it
is difficult to specify operating environment conditions precisely
in the process of building systems. A system may be used in an
environment that is different from the system building environment.
No system can work properly in an operating environment that is totally
different from the developing environment. However, some systems are
less sensitive than others regarding changes in the environment.

Building models that are insensitive to the environment changes is generally referred to as robust classification. The
robustness indicates the capability of dealing with noises and
outliers of a system. The impact of noises is twofold.
\begin{itemize}
\item noises make a data set un-learnable. No credible classifiers can
be built from the noisy data. In this case, the source of noises needs to be discovered and data needs to be cleaned. This is not a focus of our study.
\item noises make a classifier unreliable. A classifier performs well in the
known training data, but does not make reliable predictions in an
operating environment with noises. Our work is in this category. Correcting noisy operating data is an approach to make the classification results reliable. In this paper we focus on building models for an imprecise operating
environment without correcting noisy values.
\end{itemize}

Robust classification has wide applicaitons. For example,
biomedical samples may be obtained from different equipments using different processing procedures
by different methods. However, we still wish the model produced from data of one experiment to be applicable in data of another experiment. The social environment changes constantly with time. We wish
to use a model built on, say the previous five years data to the next two year
data.

A practical setting of our discussions is based on biomedical data,
where the dimension is normally very high and samples are relatively few,
for example, thousands of attributes versus tens to a few hundreds of samples. Noises are
unavoidable in biomedical experiments and can be
introduced in multiple stages, biomedical sample preparation,
experiment, data collection and data processing. Since
the number of samples is normally small, minor noises can have big impact. For example,
it has been found that one predictive gene set found from one data set performs badly on another data set~\cite{RobustGene}.

A tree based ensemble classifier is more robust than a single tree on noisy test data set. A decision tree is sensitive (or insensitive) to noises in a test data set depending on which attributes are noisy. A decision tree makes use of a small subset of attributes for classification. Inconsistencies in these attributes
between the training and test data sets will cause a significant downgrade of the classification performance of the tree. In contrast, inconsistencies outside these attributes do not affect its classification performance. When a classifier consists of a set of trees, its performance will be better than a single tree on a noisy test data set since the chance of all trees being affected by noises is smaller than the chance a single tree being affected by noises.    

A question is which tree based ensemble method is robust for classification. We demonstrate that the diversified multiple tree (DMT) approach~\cite{Hu06MDMT} is more robust than other ensemble methods, such
as AdaBoost~\cite{freund96boosting}, Bagging~\cite{breiman96bagging}, and Random Forests (RF)~\cite{Breiman01randomforests},  in real world biomedical data sets. DMT is different from those ensemble methods, which make use of a large number of weak classifiers to improve  classification accuracy. DMT utilises a small number of strong models to improve the accuracy and robustness of
a tree based ensemble classifier and has a better interpretability than the other tree based ensemble methods. Note in the previous work~\cite{Hu06MDMT},  DMT has been shown better or at least as good as the other well known tree based ensemble methods. This work will provide further evidence to support that DMT is a good choice for biomedical data classification.   

\section{Problem definition}

Let $D$ be a training data set. We build model $M$ on $D$, and apply
the model to a test data set, $D_T$. Some values in $D_T$ are
noisy. Noises make values in data deviate from their true values.

Note that noises can exist in training data set $D$ too. For
example, let us assume that $D$ is noisy and $D_T$ is not. From
model $M$'s viewpoint, $D$ is its ground truth, and relatively $D_T$
is noisy. Therefore, the assumption of $D_T$ being noisy is a
general one. Noises can be in both $D$ and $D_T$ but we offset all
inconsistencies to $D_T$. Therefore, we only consider noises in
$D_T$ in this paper. Another assumption is that noises in $D$ is not
big enough to affect the learnability of data set $D$.

Noises in $D_T$ will affect classification accuracy of model $M$ on
$D_T$. The research question is how to build a robust model $M$. The
robustness of a model is its capacity for resisting noisy values in
the test data. In other words, a more robust model will make more
accurate classifications on the noisy test data than a less robust
model.

Intuitively, an ensemble model containing more than one classification model
will be more robust than a single classification model. This is because that noisy
values are more likely to affect a single model than two or more
models simultaneously. In other words, a single model is easier to be affected by noises than an ensemble model.

Let us consider a simple vote mechanism of an ensemble model.
Each alternative model makes a classification. The final classification of the
ensemble model is the most frequent predicted class of all alternative
models. For example, if we have 5 alternative classifiers in an ensemble model where each classifier is very accurate. For a test record, even if the noise affects the classification performance of two of the five classifiers, the ensemble classifier still gives the correct classification.

The core for the robustness of an ensemble model is that its
classification is based on multiple alternative models. Assume that those
alternative models are independent from each other for being affected by
noises. In other words, noises are
uncorrelated among attributes of a data set. Assume that all models are equally accurate and have an equal probability being affected by noises. Let the probability of one model
affected by noises be $\alpha$. The probability of two models
being affected by noises simultaneously is $\alpha^2$. So, two models are less likely being affected by noises than one model. Three  models are less likely being affected by noises than two models and so forth.

The independence of models being affected by noises is the key for the robustness of an ensemble classifier. Let us assume that all attribute values have the equal chance for being affected by noises. An intuitive implementation of independent models is to build disjunct models which do not share attributes in their decision logics. As a result, disjunct alternative models are independent from each other for being affected by noises. In the next section, we will discuss a decision tree based implementation.

\section{Diversified Multiple Tree (DMT)}

A decision tree is a popular data mining method, and is a form of representation of
humanly understandable knowledge. A decision tree employs a divide and
conquer scheme for tree building. Each partition makes use of an
attribute. Normally, a decision tree makes use of a subset of all
attributes in decision. When the number of attributes is large, a decision only makes use of a small number of attributes. Other quality decision trees can be built on remaining attributes. It is easy to build a set of trees that are disjunct.

A typical decision tree construction method is
C4.5~\cite{quinlan:program}. C4.5 divides the training data into
some disjoint sub data sets based on distinct values of an
attribute, which is selected by the information gain
ratio~\cite{quinlan:program}. The sub data sets are then
simultaneously divided by other attributes recursively until each
sub data set contains instances of one class, or nearly. In
classification, a coming unclassified instance is traced down a path
from the root of the decisions tree to a leaf that contains the majority of instances of one class.
The instance is classified by the class at the leaf of the
matched path.

A decision is unreliable in an operating environment with noises since noises in some attributes may affect its performance significantly. For example, a decision tree built on the Harvard data set (to be explained in the Experiments section) makes use of only a few genes out of 11657 genes. If the data value of a gene included in the tree is noised in a future data set, the model would make wrong predictions even though other genes can help make right predictions. To make a tree model robust, multiple trees should be used.   

Multiple trees will make a tree model robust if they do not use the same attributes. A decision tree makes use of attributes for predictions explicitly, it is
easy to build disjunct decision trees to make a tree model robust. In some data sets, for
example gene expression data, the number of attributes is large, say
10,000 attributes. It can be easily to build trees on the data
set without using overlapping attributes.

A method for building disjunct multiple trees, Diversified Multiple Tree (DMT), is depicted in
Algorithm~\ref{alg:DMT}. The idea the DMT algorithm is to build a set of disjunct trees. In this
process, used attributes by a tree are knocked off. As a result, all
output trees are disjunct. The base algorithm for decision tree construction is
C4.5~\cite{quinlan:program}. The classification process is based on the
simple vote mechanism.

Since the algorithm is self-explanatory and we do not explain it here. 

\begin{algorithm}
\caption{Diversified Multiple Tree (DMT)} \label{alg:DMT} 
Training
\\Input: data set $D$, integer $k$
\\Output: ensemble model $M$
\begin{algorithmic} [1]
\STATE let $i= 0$ \STATE initiate ensemble model $M$
\WHILE{$i<k$} \STATE Build a C4.5 tree $T_i$ on $D$ \STATE remove all
attributes used in $T_i$ from $D$ \STATE let $i=i+1$ \STATE add
$M_i$ to the ensemble model $M$ \ENDWHILE \STATE output
ensemble model $M$.
\end{algorithmic}
Classifying
\\Input: a data record $r$ and an ensemble model$M$
\\Output: class label of $r$
\begin{algorithmic} [1]
\STATE $C = \emptyset$ 
\FOR {each $M_i$ in ensemble model $M$} 
\STATE let $c_i$ = classification result of $M_i$ on $r$ \STATE
add $c_i$ to $C$ \ENDFOR \STATE output the most frequent class in
$C$
\end{algorithmic}
\end{algorithm}

The complexity of a decision tree construction algorithm is linear to
the number of attributes and the size of a data set, i.e.
$O(mn\log(n))$, where $n$ is the number of data objects and $m$ is
the number of attributes. When we build $k$ trees, the complexity
becomes $O(kmn\log(n))$. The algorithm is efficient.

Alternative to the simple vote, various weighting schemes can be
used in the classification stage. One is based on the precision of
the leaf making the classification to integrate classifications of alternative tress. Since most data sets used in our
experiments are small, we use Laplace accuracy: $acc_{L} = (\#tp +
1) / (\#tp + \#fp + c)$, where $\#tp$ and $\#fp$ are the number of
true positives and false positives, and $c$ is the number of classes
in a data set. This scheme gives accurate classification high weights.
However, such weights do not consider the number of instances in a
decision leaf. To counter such a drawback, a support scheme weights the
ratio of instances at a decision leaf, i.e. support = $\#fp/n$ where $n$ is the size of a data set. We will test both schemes in our experiments.

Unlike Bagging~\cite{breiman96bagging} and Random Trees~\cite{dietterichexperimental}, DMT does not sample instances, but makes
use of different set of attributes to build diversified trees. Unlike
Random Forests and Random Trees~\cite{dietterichexperimental} which make use of attributes randomly, DMT utilises the attributes in a systematical way. Unlike Bagging, Boosting~\cite{freund96boosting}, Random Forests and Random Trees which need a large tree community to make accurate
classifications, DMT only needs a small number of high quality decision
trees.

\section{Experiments}

Experiments are conducted in two parts. The first one is to
test the robustness of DMT in comparison with other
ensemble methods. The second part is to test the effectiveness of various classification weighting schemes of DMT. All
experiments are conducted on real world high dimensional data sets.

\subsection{The robustness of DMT on noisy test data}

\begin{table}[tb]
\center
\begin{tabular}{|c|c|c|c|c|c|}
\hline Name & ~~\#Attr~~ & ~~Size~~ & ~~Classes~~ & ~~Comments~~
\\
\hline
Harvard (H) & 11657 & 156 & 139/17 & Affymetric  \\
Michigan (M) & 6357 & 96 & 86/10 & Affymetric \\
Stanford (S) & 11985 & 46 & 41/5 & cDNA  \\
\hline
\end{tabular}
\caption{Data set description}
\label{tab_datadescription}
\end{table}

Data sets come from three laboratories studying the same type of lung cancer, called Harvard~\cite{Harvarddata}, Michigan~\cite{MichiganData}, and Stanford~\cite{StandfordData}.
They have been obtained from different patient samples and from
two different Microarray platforms. There are some inconsistencies among the data sets because of different experimental environments. For one data set, another data set is noisy. We will test how DMT improves the classification accuracy when a classifier is trained by the data in one laboratory and tested on the data from another laboratory.
%
%
A brief description of three data sets is listed in Table~\ref{tab_datadescription}.

We have preprocessed the Harvard, Michigan and Stanford data sets to make
models built on them comparable. This includes removing duplicated genes in the
data sets since they correspond to different fragments of a gene and
are unable to match across labs at the name level, and removing genes
that could not match genes from another laboratories. Finally, 1963 genes
are kept in all the three data sets. Three data sets are
normalised by $z$-scores.

\begin{table}[t]
\center
\begin{tabular}{|cc|cccccc|}
\hline training & test  & ~~C4.5~~ & 7-DMT & ~~Ada~~ & ~~Bag~~ & ~~RF~~ & ~~RT~~ \\
\hline
Harvard & Michigan & \bf{99.0} & \bf{99.0} & \bf{99.0} & \bf{99.0} & 93.8 & 85.4 \\
Harvard & Stanford & 84.8 & \bf{95.7} &  89.1 & 84.8 & 91.3 & 82.6 \\
Michigan & Harvard & 96.8 & \bf{98.7} &  96.8 & \bf{98.7} & 89.1 & 89.1 \\
Michigan & Stanford & 54.3 & \bf{95.7} &  54.3 & 93.5 & 89.1 & 78.2\\
Stanford & Harvard & 80.8 & \bf{92.9} &  80.8 & 89.7 & 89.1 & 84.6 \\
Stanford & Michigan & 85.4 & \bf{94.8} &  85.4 & 85.4 & 89.6 & 86.5\\
\hline
\multicolumn{2}{|c|}{Ave} & 83.5 & \bf{96.1} & 84.2 & 92.3 & 90.3 & 84.4 \\
\hline \end{tabular} \caption{A comparison of test accuracies on data sets of different laboratories. Accuracies are in percentage. The highest accuracy in each row is highlighted. }
\label{tab_AccComp}
\end{table}

We firstly test the robustness of DMT in comparison with a single decision tree and other randomisation based ensemble methods, namely AdaBoost~\cite{freund96boosting}, Bagging~\cite{breiman96bagging}, Random Forests (RF)~\cite{Breiman01randomforests}, and Random Trees (RT)~\cite{dietterichexperimental}. The number of interactions of AdaBoost is 100. The number of trees of Bagging, Random Forests and Random Trees is 100 respectively. We have used Weka implementation of these methods~\cite{weka} for experiments. Our DMT has also been implemented as an API and plugged to Weka. The number of diversified trees is set as 3, 7, 13 and 21 respectively. In the experiment, each classifier is trained on a data set and then tested on two other data sets respectively. Because of different laboratory environments, test data sets are considered as noisy. The experiment results are summarised in Table~\ref{tab_AccComp}.

\begin{table}[tb]
\center
\begin{tabular}{|c|ccccc|}
\hline $p$-value &~~C4.5~~ & ~~Ada~~ & ~~Bag~~ & ~~RF~~ & ~~RT~~    \\
\hline 7-DMT & \bf{0.030} & \bf{0.030} & \bf{0.050} & \bf{0.016} & \bf{0.016} \\
 Ada & 0.5 & - & - & - & - \\
 Bag & \bf{0.05} & 0.14 & - & 0.28 & \bf{0.016} \\
 RF & 0.22 & - & - & - & - \\
 RT & 0.58 & - & - & - & - \\
\hline
\end{tabular}
\caption{Wilcoxon signed ranks test for various methods. The
alternative hypothesis is that the method to the left is more
accurate than one at the top. Significant test results at 95\% confidence level are highlighted.}
\label{tab_WilcoxonTest}
\end{table}
%
%
%
%

\begin{figure}[t]
\center
\begin{tabular}{c c}
    \begin{minipage}[htbp]{0.48\textwidth}
      \includegraphics[width=\textwidth, height = 4.5 cm]{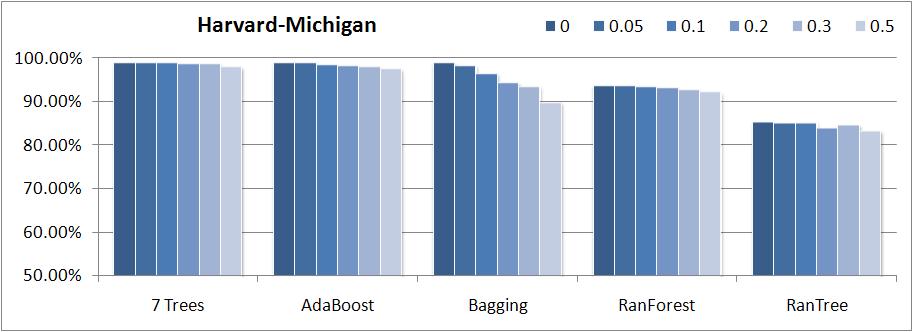}
    \end{minipage}
&
    \begin{minipage}[htbp]{0.48\textwidth}
      \includegraphics[width=\textwidth, height = 4.5 cm]{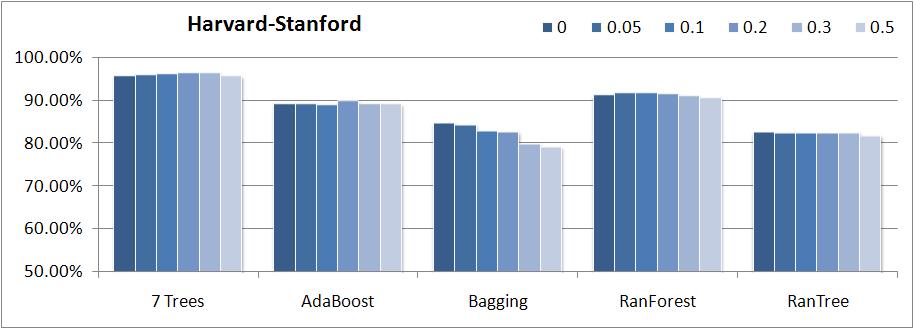}
    \end{minipage}
\\
    \begin{minipage}[htbp]{0.48\textwidth}
      \includegraphics[width=\textwidth, height = 4.5 cm]{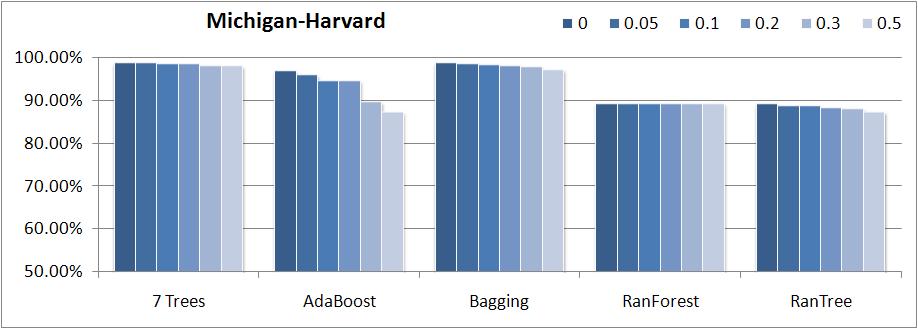}
    \end{minipage}
&
    \begin{minipage}[htbp]{0.48\textwidth}
      \includegraphics[width=\textwidth, height = 4.5 cm]{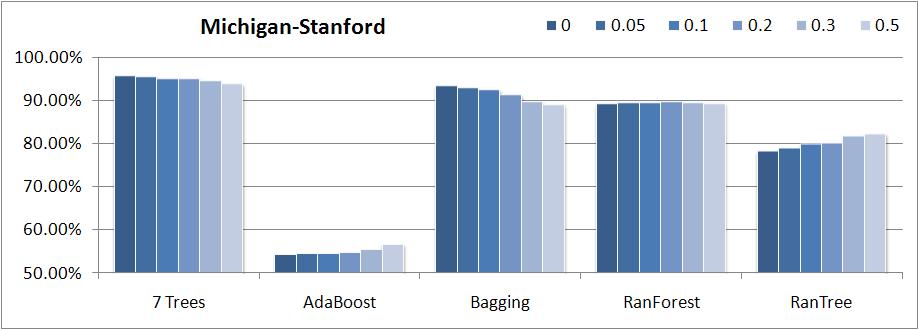}
    \end{minipage}
\\
    \begin{minipage}[htbp]{0.48\textwidth}
      \includegraphics[width=\textwidth, height = 4.5 cm]{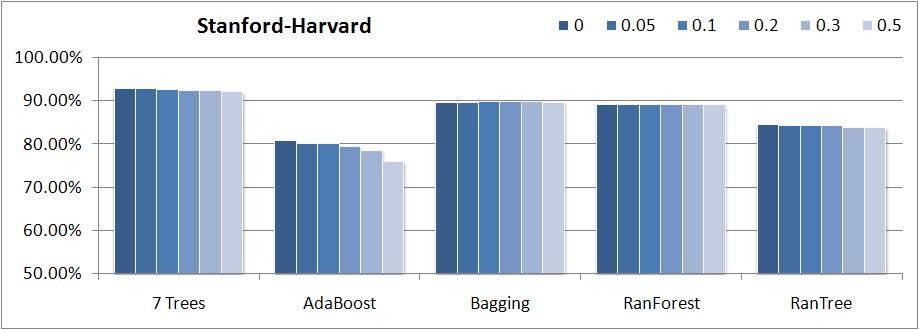}
    \end{minipage}
&
    \begin{minipage}[htbp]{0.48\textwidth}
      \includegraphics[width=\textwidth, height = 4.5 cm]{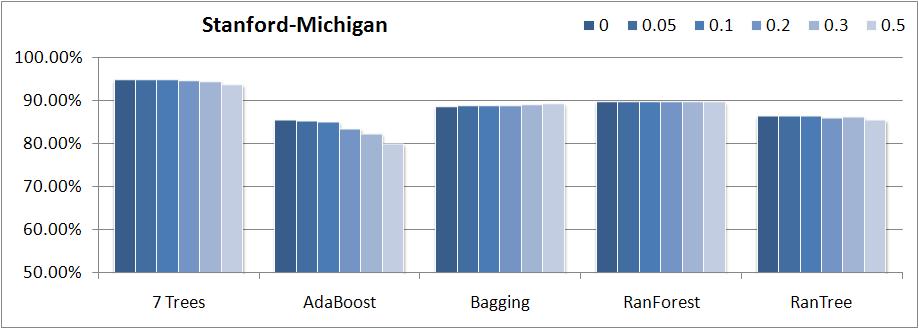}
    \end{minipage}

\end{tabular}
\caption{Test accuracy of MDTs in comparison to
various methods at different noise level} \label{fig_acc_Comp}
\end{figure}

In Table~\ref{tab_AccComp}, we see that accuracies of the single trees trained on Harvard and tested on Michigan or vice versa are higher than other single trees since both laboratories have utilised the same Microarray platform, and the inconsistencies between them are small. In contrast, inconsistencies between other training and test data set pairs are larger, and the accuracies of the single trees are lower too. DMT and all other ensemble methods improve accuracies over the single tree. The improvement of DMT is the most significant. To further confirm this conclusion, we have conducted Wilcoxon signed ranks test~\cite{BiostatTriola} to compare the performance of these methods. Dem\v{s}ar~\cite{statisticaltest06} has recommended that Wilcoxon signed ranks test is a robust non-parametric test for statistical comparisons of different classifiers. Wilcoxon signed ranks test results are listed in Table~\ref{tab_WilcoxonTest}. We see that the DMT tree is more robust than all other methods at significance level of 0.05\%. The second most robust method is Bagging, but it is not significantly more robust than AdaBoost and Random Forests. Interestingly, AdaBoost is sensitive to noises. Its performance will deteriorate greatly with the increase of noises as shown in the following experiment.

\begin{figure}[t]
\center
\begin{tabular}{c c}
    \begin{minipage}[htbp]{0.48\textwidth}
      \includegraphics[width=\textwidth, height = 4.5 cm]{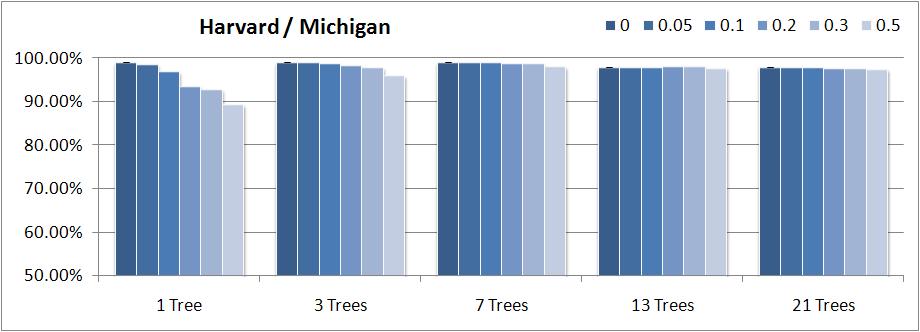}
    \end{minipage}
&
    \begin{minipage}[htbp]{0.48\textwidth}
      \includegraphics[width=\textwidth, height = 4.5 cm]{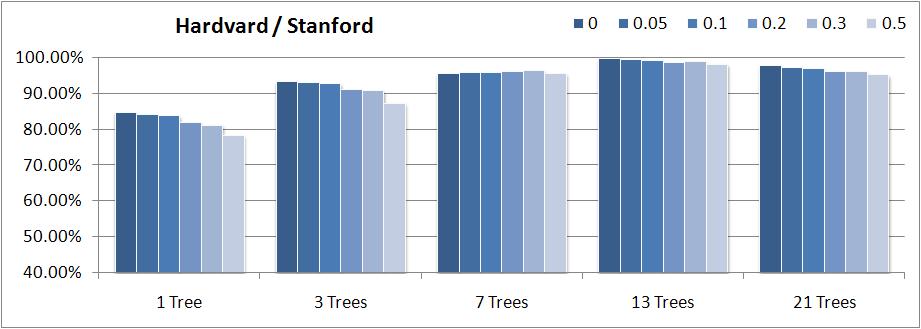}
    \end{minipage}
\\
    \begin{minipage}[htbp]{0.48\textwidth}
      \includegraphics[width=\textwidth, height = 4.5 cm]{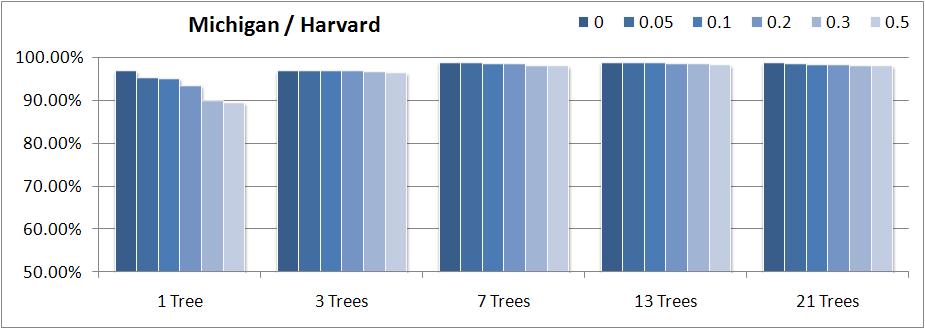}
    \end{minipage}
&
    \begin{minipage}[htbp]{0.48\textwidth}
      \includegraphics[width=\textwidth, height = 4.5 cm]{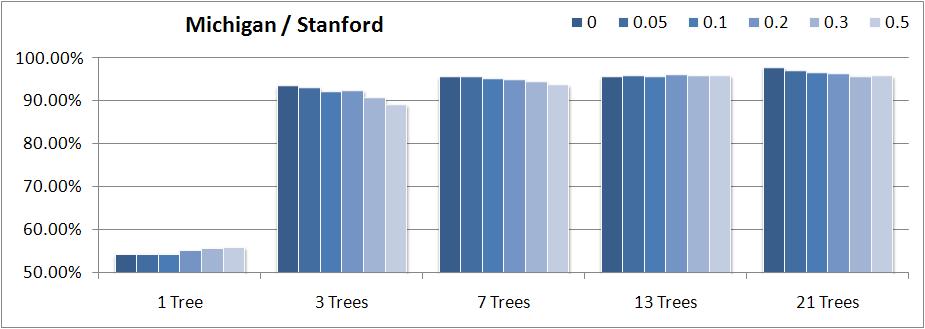}
    \end{minipage}
\\
    \begin{minipage}[htbp]{0.48\textwidth}
      \includegraphics[width=\textwidth, height = 4.5 cm]{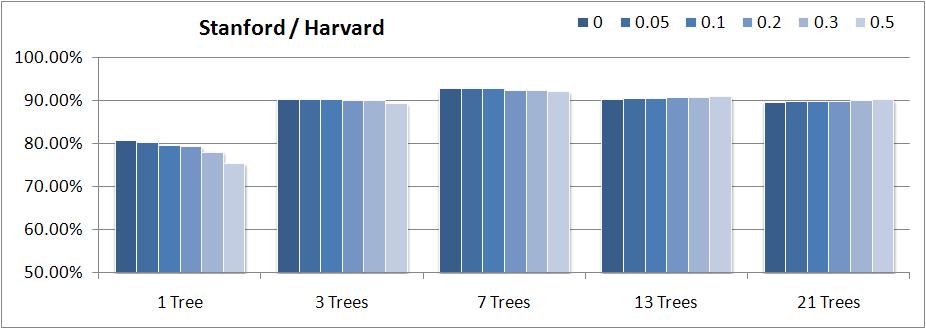}
    \end{minipage}
&
    \begin{minipage}[htbp]{0.48\textwidth}
      \includegraphics[width=\textwidth, height = 4.5 cm]{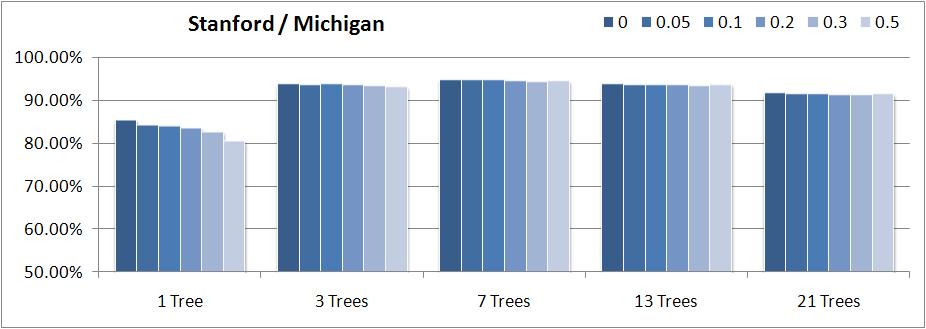}
    \end{minipage}
\end{tabular}
\caption{Test accuracy of different sizes of MDTs at
different noise levels} \label{fig_acc_DMT}
\end{figure}

To further test the robustness of DMT and various ensemble models, we have added noises on 5\%, 10\%, 20\%, 30\%, and 50\% of randomly selected attributes. The added noises
follow a $(0, \sigma)$ normal distribution, where $\sigma$ is the standard
deviation of a selected attribute, implemented according to Box
Muller transformation~\cite{BoxMullerNoise}. A reported result
is the average of 100 tests on the noised test data sets. Test
results of robustness of 7-DMT with other ensemble methods are listed in Figure~\ref{fig_acc_Comp} and the test results of robustness of various DMT trees are listed in Figure~\ref{fig_acc_DMT}.

Figure~\ref{fig_acc_Comp} shows that DMT is consistently more robust than other ensemble classifiers with the increase of added noises. A statistical test has confirmed that DMT classifiers are significantly more accurate than other ensemble methods in noise added data. Among the four ensemble methods, Random Forests and Bagging are the best. The performance of Random Forests is very stable with the increase of noises. Bagging is not as stable as Random Forests but its average accuracy is as high as that of Random Forests. The performance of Random Trees is also stable but it has a lower accuracy than Bagging and Random Forests. Adaboost performs inconsistently in noisy data and this is consistent with previous results~\cite{BostingNotWorking}. Adaboost sometimes performs badly, for example, in Michigan to Stanford pair. Considering that the number of DMT trees is small and each alternative tree is the heuristically best possible one on the complete data set, the interpretability of DMT trees is good. In contrast, a Bagging model contains 100 trees on the randomly sampled data sets and a Random Forests contains 100 random trees. The interpretability of each alternative tree
in a Bagging model and a Random Forests is not as good as an alternative tree in DMT. Therefore, DMT is more robust and has better interpretability than the other ensemble methods.

Figure~\ref{fig_acc_DMT} shows that all DMTs with different sizes are more robust than a single decision tree. Added noises do not affect their classification accuracies much. Looking at the two results of training on Harvard and testing on Michigan and vice versa, the performance of a single decision tree is quite good since both laboratories make use of the same Microarray platform and both data sets are quite consistent when there is no added noises. However, with the increase of added noises, the accuracy of a single tree reduces greatly. In contrast, DMT trees maintain a similar accuracy as for that of no noise added data. Added noises make slight difference in classification accuracies for DMTs, and hence DMT trees are robust.

The robustness of DMT trees does not necessarily need a large number of alternative trees. In overall, 7-DMT and 13-DMT  are the best among all DMTs. This has been typified in results of Stanford to Harvard and Standard to Michigan. Theoretically, a large number of diversified tees will increase the robustness. However, there may not be the same high quality trees as the first few trees on a data set when the number of trees increases. Low quality trees reduce classification accuracies. Therefore, the number of DMT trees should not be large.

\subsection{A counter example}

\begin{figure}[h]
\center
\includegraphics[width=0.5\textwidth, height = 4.5 cm]{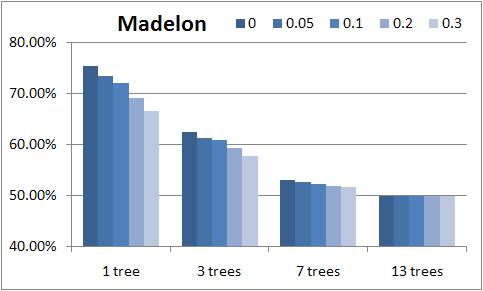}
\caption{DMT does not work on Madelon data set.} \label{fig_Madelon}
\end{figure}

A base of the robustness of DMT trees is that a data set supports a number of quality alternative trees. Thus is true for many high dimensional biomedical data sets. When this is not true, we should expect that DMT does not work well. We have identified an example and show it in Figure~\ref{fig_Madelon}.

The Madelon data set from UCI ML repository~\cite{Asuncion+Newman:2007} is an artificial dataset containing
data points grouped in 32 clusters placed on the vertices of a five
dimensional hypercube and randomly labeled as two classes. The data
points are described by 500 features (attributes). The training data
set contains 2000 data points (1000 in each class) and the test data set
contains 1800 (900 in each class) data points. The performance of MDT is bad
in all aspects. The accuracy of un-noised and noised test data of DMT
is lower than a single decision tree.

Let us compare the size of trees in the MDT classifiers of Madelon and
of Harvard in Table~\ref{Tab_treeszie}. We see that the tree sizes for Madelon
are large in the first four trees, and are very small after the fourth tree. Firstly, a large tree indicates that it is difficult to build a good classifier on the training data. The
 size increases in the subsequent trees and this shows that
the classification becomes even more difficult in the remaining attributes. Secondly, a tree of size one classifies a test instance to the largest distributed class. In this data set, the classification of a tree of size one is a random choice since both classes are equally distributed. With the above knowledge, we can easily understand Figure~\ref{fig_Madelon}. The first tree is the most accurate one and all subsequent trees are less accurate than their previous tree in the first four trees. Therefore, all DMT classifiers are less accurate than the first decision tree. Further, a large tree is more likely affected by noises than a small tree. DMT performance deteriorates with the increase of noises. For DMTs whose sizes are greater than 8, their classification accuracy is around 50\%.

In contrast, trees built with the Harvard data set are consistently small.
This means that many alternative trees are very good for the classification of Harvard data set. Many quality small trees make DMT work well.
However, when good classification attributes are
inadequate, alternative trees are significantly worse than the
first tree. As a result, DMT does not work for the Madelon data set.

\begin{table}[tb]
\center
\begin{tabular}{|c|cccccccc|}
\hline Tree size & ~~1st~~ & ~~2nd~~ & ~~3rd~~ & ~~4th~~ & ~~5th~~ & ~~7th~~ & ~~13th~~ & ~~21th~~ \\
\hline Madelon & 259 & 379 & 409 & 437 & 1 & 1 & 1 & 1 \\
Harvard & 3 & 3 & 3 & 5 & 7 & 7 &  5 & 9 \\
\hline
\end{tabular} 
\caption{Tree size of DMT on Madelon data set in comparison with Harvard data set.}
\label{Tab_treeszie}
\end{table}

\subsection{Variations of DMT}

We have discussed a number of possible variations for the construction of DMT classifiers: simple vote, weighted vote with Laplace accuracy at a predictive leaf, and weighted vote with the support at
a predictive leaf.  

\begin{figure}[t]
\center
\includegraphics[width=0.5\textwidth, height = 4.5 cm]{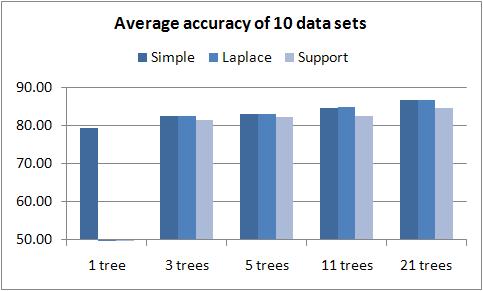}
\caption{Cross validation accuracy of MDTs of
ten data sets} \label{fig_acc_CompNoNoise}
\end{figure}

We test these variations on the 10 data sets with 3, 5,
11 and 21 trees. The 10 data sets have obtained from Kent Ridge~\cite{KentRidge} (the first six) and research literature~\cite{centralNeverSys,prostateCancer,BCTamoxifen,Liver-tumor} (the last four). A brief description of the data sets is listed in Table~\ref{tab_description2}. The 10 cross validation accuracies of various variations are listed in Figure~\ref{fig_acc_CompNoNoise}.

\begin{table}[t]
\center
\begin{tabular}{|c|ccc|}
\hline Name & ~~\#Att~~ & ~~Size~~ & ~~Class~~ \\
\hline Breast Cancer & 24481 & 97 & 51/46  \\
Lung Cancer & 12533 & 181 & 150/31  \\
DLBCL & 4026 & 47 & 24/23   \\
AllAML & 7129 & 72 & 47/25 \\
Colon tumor & 2000 & 62 & 40/22  \\
Ovarian & 15154 & 253 & 162/91  \\
Central Nervous Syst & 7129 & 60 & 39/21 \\
Prostate Cancer & 12600 & 136 & 102/34 \\
BC Tamoxifen & 21939 & 120 & 60/60 \\
Liver tumor & 22686 & 180 & 105/75 \\
\hline
\end{tabular}
\caption{Data set description}
\label{tab_description2}
\end{table}

From Figure~\ref{fig_acc_CompNoNoise}, we see that the difference between simple vote and the Laplace weighted vote is marginal, and that the support weighted vote is worse than both. Therefore, simple vote is a simple yet accurate choice. In addition, all DMT classifiers in this experiment are significantly more accurate than a single decision tree without noises.

\section{Related work}

Our method is an ensemble method and in this section we review other ensemble classification methods used in this paper. 

Bagging~\cite{breiman96bagging}, Boosting~\cite{freund96experiments}, Random Forests~\cite{Breiman01randomforests} and Random Trees~\cite{dietterichexperimental} are four major randomisation based ensemble
classification methods in the machine learning field. We present a brief discussions for each in the following.

Bagging was proposed by Breiman~\cite{breiman96bagging}.
Bagging uses a bootstrap technique to re-sample a training data
set. Bootstrap sampling is random sampling with replacement.
A set of alternative trees are built on a set of re-sampled
data sets. Each tree will gives a predicted class to
a coming test instance. The final predicted class for the test instance
is determined by the majority predicted class by all alternative trees.

Boosting method was first developed by Freund and
Schapire~\cite{freund96experiments}. Boosting trains a
sequence of classifiers on a set of data sets with different
distribution ratios. The first classifier is constructed from the
original data set where every record has an equal distribution ratio
of 1. In the following training data sets, the distribution ratios
are assigned differently among records. The distribution ratio of a
record is reduced if the record has been correctly classified, and
is increased otherwise. A weighted voting method is used in the
committee decision.

Random Forests was proposed by
Breiman~\cite{Breiman01randomforests}. This method combines the Bagging
and Random Subspace methods~\cite{Ho98therandom}. A random forests classifier consists of a
number of randomly generated trees. Each tree is constructed from a
bootstrap sample of the original data set. At each node of a tree,
the data partition attribute is selected from a subset of randomly
chosen attributes.

Random Trees was proposed by Dietterich~\cite{dietterichexperimental}. In random tree construction, at each node in the decision tree one randomly selected attribute from the twenty best tests is selected to partition data at the node. With continuous attributes, it also produces twenty best splits of attributes, and the one used is randomly selected from the top 20.

All these randomisation-based ensemble methods are based on a large set of weak learners to improve classification accuracy of tree based classifiers. Their performance on un-noised data has been comprehensively evaluated at~\cite{EnsembleTReeCompTPAMI2007}. In contrast, DMT~\cite{Hu06MDMT} makes use of a small set of disjunct strong trees and it has been shown to outperform AdaBoost, Bagging, and Random Forests for microarray data classification. In this paper, it has been shown to outperform AdaBoost, Bagging, Random Forests and Random Trees on noisy test data sets.


\section{Conclusions}

This paper has demonstrated that a Diversified Multiple Trees (DMT) approach is able to classify noisy test data that is deviated from the training data, with better performance than some widely used ensemble methods. We have tested DMT on three real world biomedical data sets from different laboratories in comparison with four benchmark ensemble classifiers, and have experimentally shown that DMT is significantly more robust than the other benchmark ensemble classifiers on noisy test data. We have also discussed a limitation of DMT when a training data set does not support many simple and quality decision trees. We have further demonstrated that DMT is a simple yet effective design among a number of possible variations. DMT makes use of a small number of quality trees to tolerate noises in test data. It is more robust than other large ensemble classifiers and has better interpretability. It is promising in many real world applications where data dimension is high and noises are present.

\section*{Software tool}
The software tool is available at \url{http://nugget.unisa.edu.au/jiuyong/MDMT/MDMT.html}.

\bibliographystyle{plain}
\bibliography{../../bioinfo,../../emsemble,../../robust,../../reference}

\end{document}